\documentclass[conference]{IEEEtran}
\IEEEoverridecommandlockouts
\usepackage{cite}
\usepackage{amsmath,amssymb,amsfonts}
\usepackage{algorithmic}
\usepackage{graphicx}
\usepackage{textcomp}
\usepackage{xcolor}

\def\BibTeX{{\rm B\kern-.05em{\sc i\kern-.025em b}\kern-.08em
    T\kern-.1667em\lower.7ex\hbox{E}\kern-.125emX}}

\usepackage{tikz}

\newcommand\copyrighttext{%
  \footnotesize \textcopyright 2025 IEEE. Personal use of this material is permitted. Permission from IEEE must be obtained for all other uses, in any current or future media, including reprinting/republishing this material for advertising or promotional purposes, creating new collective works, for resale or redistribution to servers or lists, or reuse of any copyrighted
component of this work in other works.

This work has been accepted for oral presentation at the IEEE ICE 2025
}
\newcommand\copyrightnotice{%
\begin{tikzpicture}[remember picture,overlay]
\node[anchor=south,yshift=10pt] at (current page.south) {\fbox{\parbox{\dimexpr\textwidth-\fboxsep-\fboxrule\relax}{\copyrighttext}}};
\end{tikzpicture}%
}

\begin{document}
\title{Personalized Mental State Evaluation in Human-Robot Interaction using Federated Learning}

\author{
    \IEEEauthorblockN{Andrea Bussolan}
    \IEEEauthorblockA{
        \textit{ARM-Lab} \\
        \textit{University of Applied Sciences} \\ 
        \textit{of Southern Switzerland (SUPSI)} \\
        Lugano, Switzerland \\
        andrea.bussolan@supsi.ch
    }
    \\
        \IEEEauthorblockN{Stefano Baraldo}
    \IEEEauthorblockA{
        \textit{ARM-Lab} \\
        \textit{University of Applied Sciences} \\ 
        \textit{of Southern Switzerland (SUPSI)} \\
        Lugano, Switzerland \\
        stefano.baraldo@supsi.ch
    }
    \and
    \IEEEauthorblockN{Oliver Avram}
    \IEEEauthorblockA{
        \textit{ARM-Lab} \\
        \textit{University of Applied Sciences} \\ 
        \textit{of Southern Switzerland (SUPSI)} \\
        Lugano, Switzerland \\
        oliver.avram@supsi.ch
    }
    \\
    \IEEEauthorblockN{Anna Valente}
    \IEEEauthorblockA{
        \textit{ARM-Lab} \\
        \textit{University of Applied Sciences} \\ 
        \textit{of Southern Switzerland (SUPSI)} \\
        Lugano, Switzerland \\
        anna.valente@supsi.ch
    }
    \and
    \IEEEauthorblockN{Andrea Pignata}
    \IEEEauthorblockA{
        \textit{DAUIN Department}\\
        \textit{Politecnico di Torino} \\ 
        Torino, Italy \\
        andrea\_pignata@polito.it
    } 
    \and
    \IEEEauthorblockN{Gianvito Urgese}
    \IEEEauthorblockA{
        \textit{DIST Department} \\
        \textit{Politecnico di Torino} \\ 
        Torino, Italy \\
        gianvito.urgese@polito.it
    } 
}

\maketitle
\copyrightnotice
\begin{abstract}
With the advent of Industry 5.0, manufacturers are increasingly prioritizing worker well-being alongside mass customization. Stress-aware Human-Robot Collaboration (HRC) plays a crucial role in this paradigm, where robots must adapt their behavior to human mental states to improve collaboration fluency and safety. This paper presents a novel framework that integrates Federated Learning (FL) to enable personalized mental state evaluation while preserving user privacy. By leveraging physiological signals, including EEG, ECG, EDA, EMG, and respiration, a multimodal model predicts an operator's stress level, facilitating real-time robot adaptation. The FL-based approach allows distributed on-device training, ensuring data confidentiality while improving model generalization and individual customization. Results demonstrate that the deployment of an FL approach results in a global model with performance in stress prediction accuracy comparable to a centralized training approach. Moreover, FL allows for enhancing personalization, thereby optimizing human-robot interaction in industrial settings, while preserving data privacy. The proposed framework advances privacy-preserving, adaptive robotics to enhance workforce well-being in smart manufacturing.
\end{abstract}


\section{Introduction}

With the rise of Industry 5.0, manufacturers are increasingly prioritizing worker well-being while addressing the growing demand for mass customization. In this context, stress-aware human-robot collaboration (HRC) is emerging as a strategic asset rather than a mere optional upgrade \cite{10.3389/frobt.2022.813907}. Recent research highlights a wide range of benefits, including improved operational efficiency and safety \cite{info15100640}, alleviation of labor shortages \cite{buss2023}, meeting personalization demands \cite{10.1115/1.4062430}, and cost reductions in training through AI-driven adaptation \cite{AdekolaAdebayo2024RoboticsIM}.

In particular, personalization has emerged as a cornerstone of effective HRC, enabling robots to adapt their behaviors, communication styles, and decision-making processes to individual users’ preferences, goals, and contexts \cite{buss2023} \cite{yang_impact_2024}. To achieve this level of adaptation, it is crucial to consider also the user's mental state, as understanding factors such as stress, cognitive load, or emotional conditions allows robots to respond more appropriately and maintain smooth, natural interactions. However, meeting the personalization requirements in HRC remains a complex challenge, particularly due to privacy constraints, data heterogeneity, and real-time adaptation requirements associated with learning from diverse user interactions.

\textit{Federated Learning} (FL), first introduced by \cite{mcmahan_2017}, has been proposed as a transformative paradigm in the realm of artificial intelligence, enabling collaborative model training across decentralized devices without compromising data privacy \cite{kairouz_2021}. While FL mitigates many privacy and scalability concerns, it only addresses part of the personalization challenge \cite{checker2024federatedlearningsociallyappropriate}. A key limitation of standard FL approaches is that even after aggregation, the global model may not fully capture highly specific user preferences or behaviors \cite{10309661}. This limitation arises because FL primarily optimizes for generalization across a distributed set of users rather than precise adaptation to any single individual \cite{10.5555/3495724.3496024}. 

Despite ongoing challenges in multi-modal sensor fusion \cite{bussolan2024} and workforce acceptance \cite{10.1145/3399433}, advancements such as the deployment of advanced AI models on resource-constrained edge devices\cite{avram_advancing_2024}, localized data processing \cite{XIANJIA2021135}, adaptation to new users and contexts \cite{10309661} and federated learning as distributed learning approach for privacy preservation \cite{GAMBOAMONTERO2023118510} present compelling value propositions for the industry.

In this work, we address the need for behavior personalization in industrial HRC by proposing an FL framework that leverages individual biometric data to enhance robotic adaptation policies. Simultaneously, we ensure privacy protection by processing this data on a personal device.

The remainder of this paper is organized as follows. Section \ref{sec: sota} presents the related works; Section \ref{sec: framework} introduces the personalization framework proposed in this work; Section \ref{sec: mental} explains the developed mental state model and its inputs and outputs; while Section \ref{sec: sw} analyze the hardware infrastructure required for our proposed framework; Section \ref{sec: exp} and \ref{sec: res} describe the experimental setup used to test of proposed approach and the obtained results. Finally, Section \ref{sec: disc} and \ref{sec: concl} discuss our solution limitations and benefits and conclude the work.

\section{Related Works} \label{sec: sota}

While many studies recognize the importance of personalization, a significant gap exists in the application of consistent personalization methods leveraging mental state assessment within FL frameworks \cite{10.3389/fdgth.2024.1495999}. 

FL is gaining particular interest in the \textit{Affective Computing} field due to its subjective, personal, and private data nature. For example,  \cite{somandepalli_federated_2022} investigates the application of FL in affective computing tasks, focusing on two paradigms: user-as-client and rater-as-client. It evaluates the performance of the FedAvg algorithm for classifying self-reported emotional experiences and perceptual judgments across various datasets. 
In \cite{tsouvalas_privacy-preserving_2022}, the authors presented a novel Speech Emotion Recognition (SER) approach that prioritizes user privacy through FL and semi-supervised learning techniques. This work addresses the challenge of limited labeled data by utilizing labeled and unlabeled data on user devices, enhancing model performance without compromising privacy. The authors of \cite{zhao_multimodal_2022} propose a multimodal and semi-supervised FL framework designed to handle local data from various modalities in Internet-of-Things (IoT) environments. They introduced a multimodal FedAvg algorithm that aggregates models from unimodal and multimodal clients, enhancing classification performance. In \cite{fenoglio_federated_2023}, the researchers present a framework for accurate Cognitive Workload (CW) estimation using Federated Learning, addressing challenges like client heterogeneity, data limitation, and Out-of-Distribution (OoD) clients. 
It employs a context-based STRNet architecture for joint learning across heterogeneous datasets, specifically COLET and ADABase, enhancing model generalizability.

The importance of guaranteeing the human operator's well-being is one of the key aspects in the Industry 5.0 paradigm \cite{eu5.0}. This can be achieved with the deployment of cobots equipped with the ability to assess the human colleague's mental state \cite{villani_promoting_2022}. Given the assessment of the mental state, these cobots can adapt their behavior to improve the operator's well-being. Different examples of behavior adaptation can be found in the literature. The work of \cite{lagomarsino_pro-mind_2024} introduces a novel human-in-the-loop framework designed to optimize robot trajectories by leveraging the ECG data of human co-workers. It dynamically adjusts safety zones and robot paths based on estimated human attention and mental effort, enhancing comfort and optimal stopping conditions. In \cite{landiRelievingOperatorsWorkload2018}, they focused on analyzing the operator's mental workload while teleoperating with an industrial robot. Based on measured HRV metrics, they implemented a feedback loop to ease the maneuvering workload through virtual fixtures. 

All these approaches can benefit from a reliable estimation of the human psycho-physical state to trigger the robot's behaviors. Numerous studies focus on the effects of psychological stress and cognitive load on physiological measurements, which can serve as a reliable and non-invasive identification method leveraging machine learning models. However, the robust performance required in an industrial human-robot collaboration setting can benefit from the use of personalization of the prediction model. In \cite{patel_mental_2018}, the authors mention that current mental state assessment approaches rely on generic population-level ranges, which are insufficient for accurately predicting and enhancing an individual's near-term task performance. By generating personalized models of mental fatigue, stress, and attention, the researchers demonstrate improved prediction accuracy and reliability, which can be applied to real-time human-machine interfaces to optimize task performance.

While significant strides have been made in the above areas, the application of federated learning to trigger adapted robot behavior based on personalized human state evaluation represents a novel and relatively unexplored research direction. 

\section{Proposed framework} \label{sec: framework}

Building on these premises, we propose a novel framework facilitating the deployment of personalized mental state prediction models to improve human-robot collaboration in industrial settings. The key contributions of this work are summarized below. 

\begin{itemize}
    \item \textbf{Multimodal Physiological Signal Integration and Synchronization}: our system synchronizes and processes Electroencephalogram (EEG), Electrocardiogram (ECG), Electrodermal Activity (EDA), Electromyography (EMG), and Respiration (RESP) which are used as inputs in a multimodal deep learning model to assess human stress levels.
    \item \textbf{Personalized Mental State Estimation}: we implement a local fine-tuning strategy that adapts the global model to individual users, improving stress prediction accuracy over global models.
    \item \textbf{Privacy-preserving model improvement}: thanks to the Federated Learning approach, we continuously update the baseline mental state estimation model while avoiding transmitting raw biometric data, thus ensuring high privacy standards.
    \item \textbf{Industrial HRC Application}: the proposed system is tested in an aircraft engine assembly task, demonstrating its ability to enhance adaptive robot behavior in real-world industrial scenarios.
\end{itemize}

The framework was developed within project \textit{Fluently}, which proposed the \textit{Robo-Gym} concept, i.e., a mutual learning environment for HRC: while human operators learn how to work with AI-powered cobots, the cobots learn how to best support their fellow human teammates. Such support is also provided by estimating the human operator's mental state, which is the key enabler for empathic behavior adaptation.

\begin{figure}
    \centering
    \includegraphics[width=0.75\linewidth]{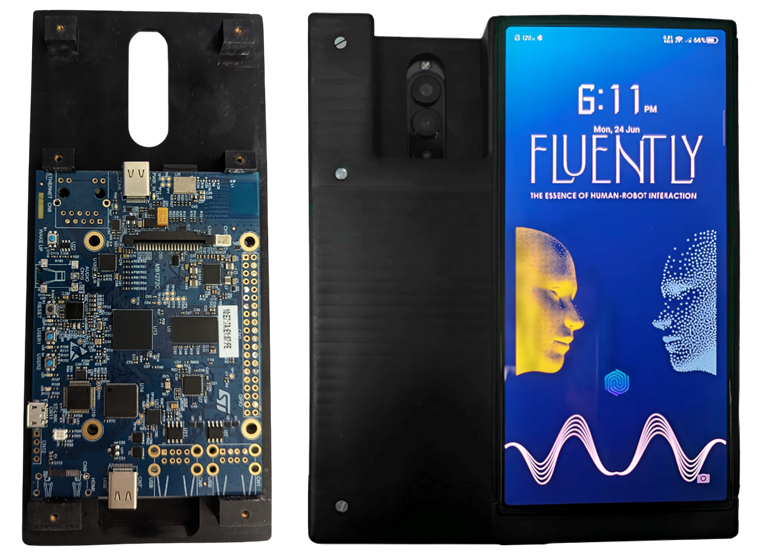}
    \caption{H-F device with a custom-designed cover to allow the compact integration of the two devices.}
    \label{fig: hf}
\end{figure}

A detailed understanding of human states and their real-world applicability in scenarios where robots must learn and adapt to highly variable contexts, such as human behaviors and preferences, was facilitated by the implementation of an intelligent communication interface. This interface consists of two devices: a mobile device used by the human operator (\textit{Human-Fluently}, briefly H-F) and a more powerful machine that interfaces with the robot (R-F). Briefly, the H-F (1) captures audio and physiological information from a human operator, (2) processes the data onboard (speech-to-text and mental state estimation), and (3) transfers the input to the R-F device. The R-F (1) interprets the received human data, (2) translates the command to robot instructions and sends it to the robot controller, and (3) collects outputs from the robot, interprets, and converts them into natural language, which is reproduced by an artificial voice synthesizer, and displaying the real-time outputs on the device.
The H-F system integrates two primary components: the STM32MP157 board with Linux real-time operating system, which gathers data from biophysical sensors, and a mobile phone featuring Qualcomm's Snapdragon 8Gen2 SoC using Android, which provides the necessary computational power for AI functionalities (i.e., on-device training, automatic speech recognition, and mental state prediction).
Both components are connected via a Y-shaped USB-C cable. This setup allows the STM32MP157 board to be powered by the smartphone's battery, with an option to connect to an external power source if the smartphone's battery needs recharging. The H-F device is represented in Fig. \ref{fig: hf}

\begin{figure}[t]
    \centering
    \includegraphics[height=5cm, keepaspectratio]{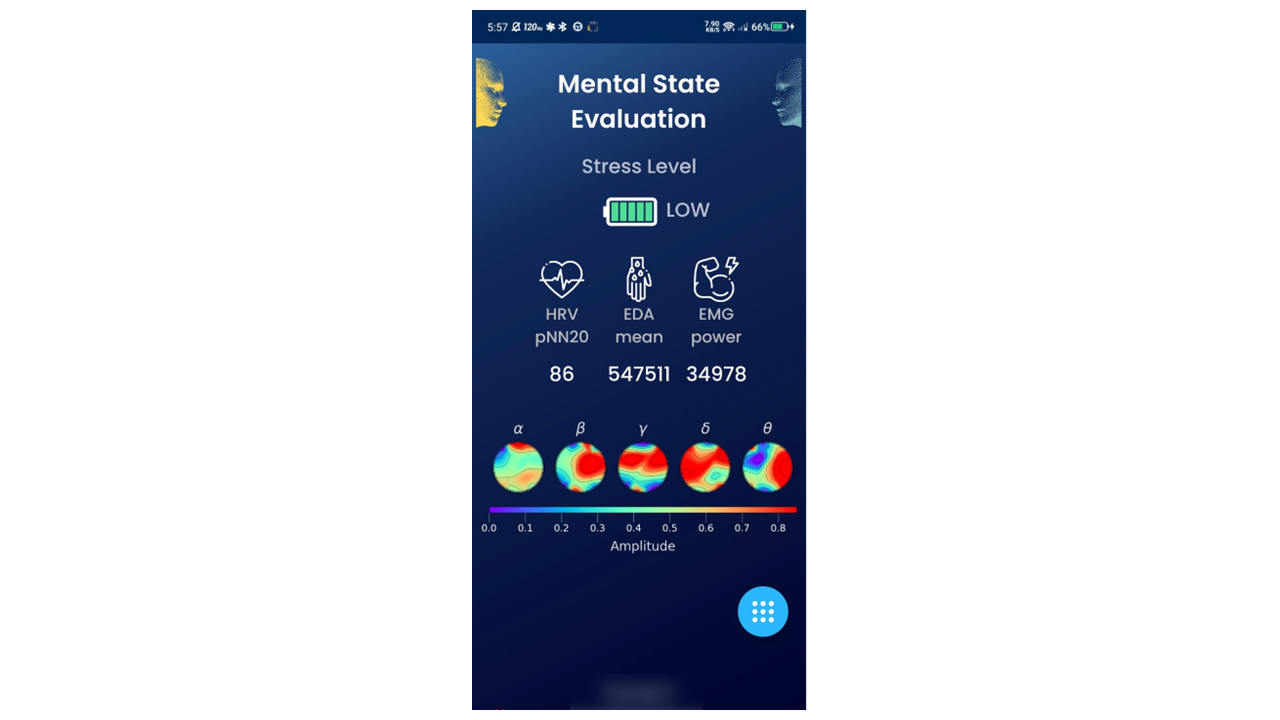}
    \caption{Mental state real-time inference in the H-F application.}
    \label{fig: screen}
\end{figure}

Several devices are available to collect data related to the human operator, specifically the Bitbrain Versatile Bio\footnote{https://www.bitbrain.com/neurotechnology-products/biosignals/versatile-bio}, which collects multiple biosignals using adhesive Ag/AgCl electrodes, and the Bitbrain Diadem\footnote{https://www.bitbrain.com/neurotechnology-products/dry-eeg/diadem}, which is a wearable dry-EEG with 12 channels. 
The mental state evaluation model is deployed on the H-F device, where data processed by the STM32MP157 board are stored locally. In particular, the H-F device handles the on-device model updates, the FL client module, the data storage, and model inference. After processing the audio stream and the physiological data, H-F outputs the operator state and its vocal instructions. The H-F device shows the real-time output, as displayed in Fig. \ref{fig: screen}, with relevant metrics computed from the physiological signals. These data are sent via Wi-Fi to the R-F, using the MQTT communication protocol, and converted into ROS2 topics, making them available to other interested applications deployed inside the R-F device.

\begin{figure}
    \centering
    \includegraphics[width=0.8\linewidth]{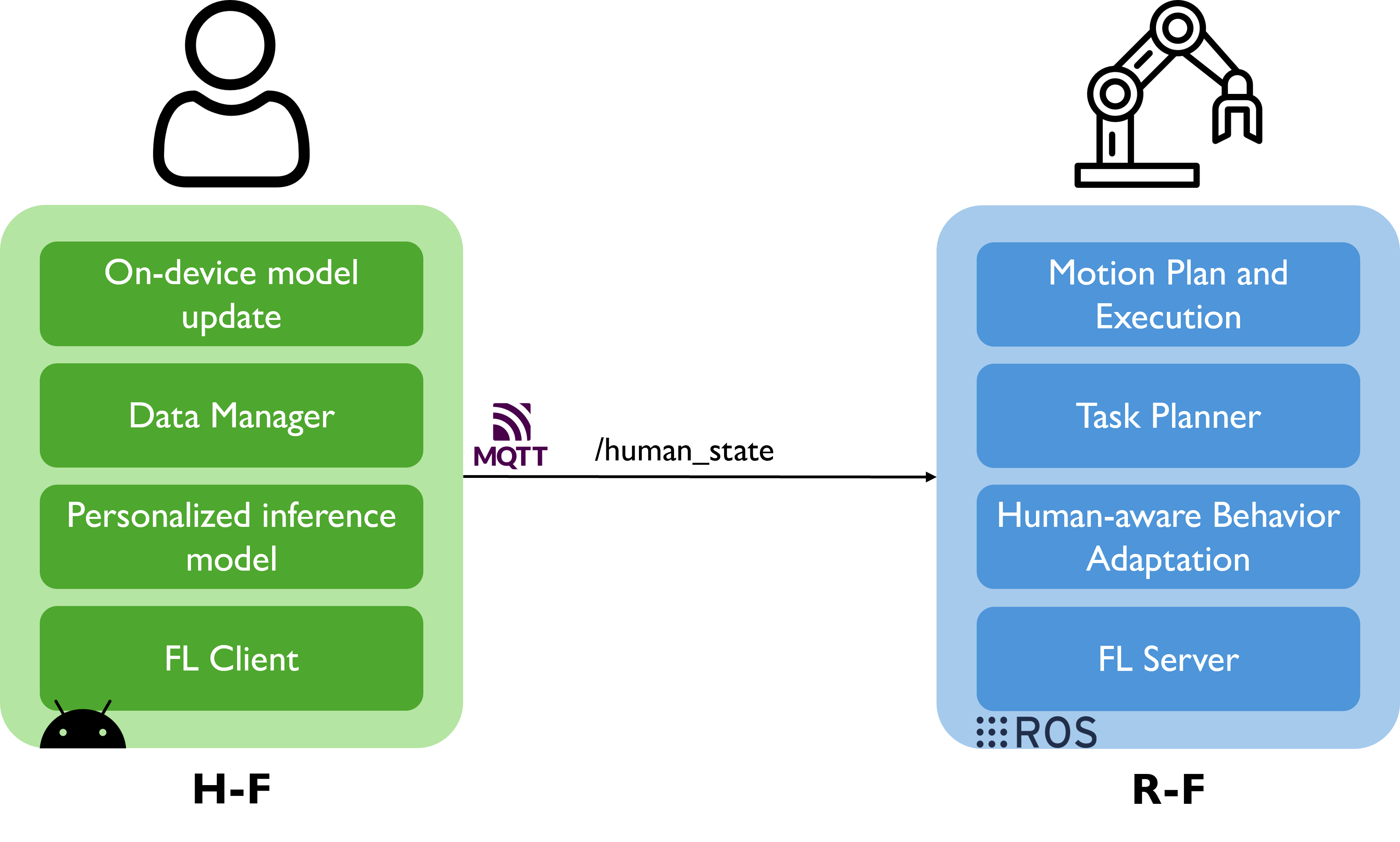}
    \caption{Communication between the various modules deployed on the H-F and R-F devices.}
    \label{fig: fl_modules}
\end{figure}

After receiving the human state, the R-F processes this information through the human-aware behavior adaptation component, which is a ROS2 node part of the overall robot architecture. This node processes the human state in real time and matches with the operator preferences and stress thresholds to adjust the robot behavior accordingly, enabling adaptation both at the task planning level and motion execution level. Specifically, the robot can activate its deliberative capabilities. For example, the robot can dynamically reallocate tasks to balance the workload in collaborative scenarios or modify trajectory speed and shape \cite{zanchettin2013} to reduce cognitive or physical load when high stress or fatigue is detected \cite{valente2022, Avram2022}. The main modules deployed in each device are represented in Fig. \ref{fig: fl_modules}. 

\begin{figure*}[t]
    \centering
    \includegraphics[width=\linewidth]{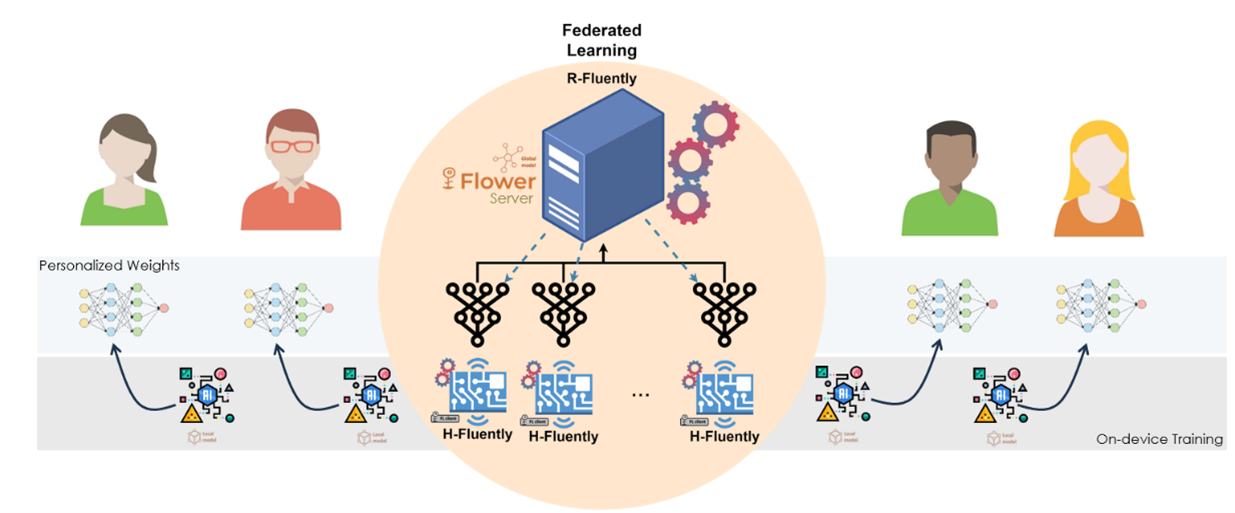}
    \caption{Schematic representation of the Federated Learning framework developed.}
    \label{fig: framework}
\end{figure*}

Leveraging the Flower framework \cite{beutel_2020}, on-device training is performed, and the weights of the local personalized models are subsequently shared with the R-F server to update the global model. A representation of the proposed framework is displayed in Fig. \ref{fig: framework}. The Flower framework facilitates federated learning by enabling collaborative model training across decentralized devices while preserving data privacy. The proposed framework consists of the following key components:
\begin{itemize}
    \item \textbf{On-Device Training}. Each H-F device trains a mental state evaluation model locally using the data collected and processed by the H-F. This ensures that raw data never leaves the device, maintaining user privacy and data security.
    \item \textbf{Global model and weight aggregation}. The locally trained model weights are then securely transmitted to the R-F server. The Flower server orchestrates this process, aggregating the weights from multiple devices/clients to update the global model. This aggregation is performed using the Federated Averaging (FedAvg) strategy, which involves iteratively collecting weights from a subset of clients, averaging these weights, and updating the global model. The process is repeated over several rounds until the global model converges to an optimal state.
    \item \textbf{Privacy}. Security and privacy of the data are preserved using this framework. The global model training process requires the exclusive sharing of the client’s model weights and never asks for raw personal data.
    \item \textbf{Personalization}. Once the global model sends its updates to the clients, it is fine-tuned on the specific subject data.
\end{itemize}

\section{Data streaming and synchronization} \label{sec: sw}

As outlined in Section \ref{sec: framework}, the evaluation of mental state is facilitated by Bitbrain devices, which are equipped with heterogeneous biological sensors. These devices generate a continuous stream of raw data, transmitted via Bluetooth Classic, enabling subsequent processing by the receiver system. The integration of these devices into the H-F system has been achieved through the establishment of a dataflow that incorporates the STM32MP157 Single Board Computer (SBC) from ST Microelectronics for data acquisition and synchronization, in conjunction with the Android-based device dedicated to Machine Learning tasks.


The STM32MP157 board provides comprehensive support for the acquisition of biological signals, facilitated by its incorporation of Linux and Bluetooth functionality. It has been equipped with a customized OpenEmbedded Linux distribution\footnote{https://www.openembedded.org/wiki/Main\_Page}, incorporating support for the Bluetooth protocol utilized by BitBrain devices, the MQTT protocol for communication with external devices, and the Bitbrain Software Development Kit (SDK)\footnote{https://www.bitbrain.com/neurotechnology-products/software/programming-tools }. The SDK offers a C++ interface to the sensors, ensuring compatibility with both Linux and Windows operating systems. Finally, a client has been written in C++ language to receive and process the biological signals.   

\begin{figure}
    \centering
    \includegraphics[width=0.99\linewidth]{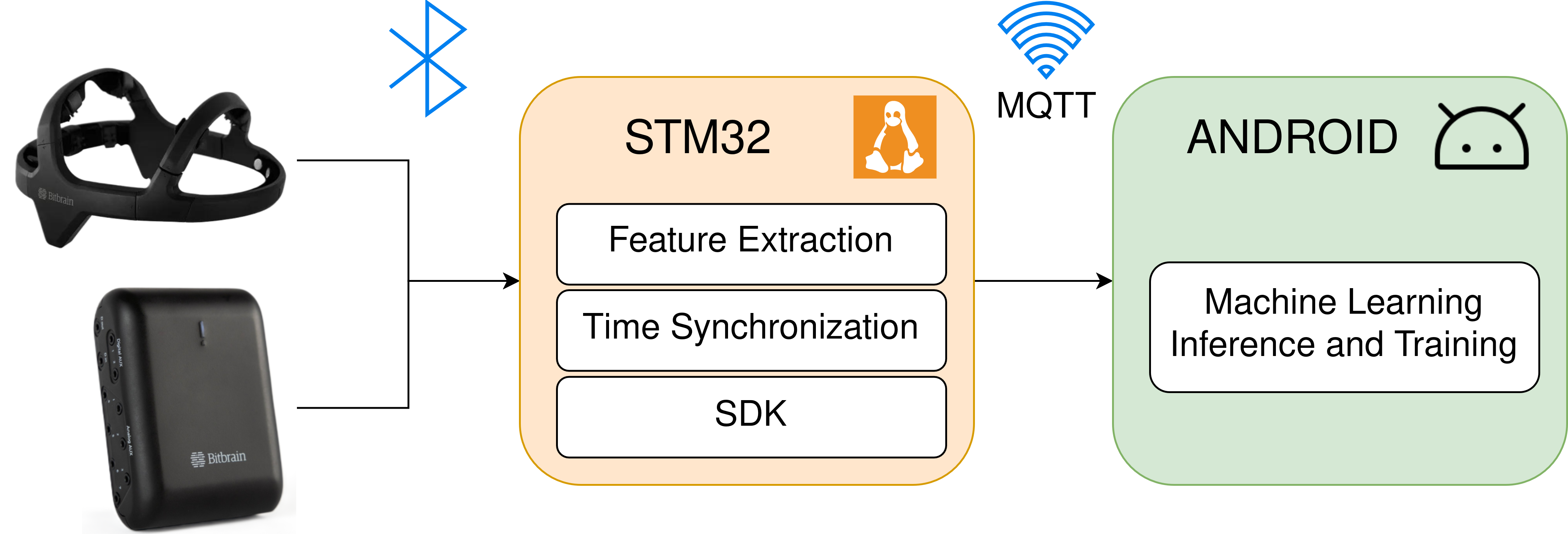}
    \caption{Data processing BitBrain devices - HF}
    \label{fig:data-processing-bitbrain}
\end{figure}

As illustrated in Fig. \ref{fig:data-processing-bitbrain}, the STM32MP157 board receives the signals from Bitbrain products via Bluetooth and the custom client software. At this stage, the SDK module is used to establish the connection with the sensors, configure them and acquire the sampled values. Once the data is received, a client submodule, provided by Bitbrain, performs lightweight feature extraction operations on the EEG signal, computing values for the alpha, beta, delta, gamma, and theta brain waves. The user can specify the biological signals to be sampled and the required post-processing steps by compiling a JSON-formatted configuration file.

Finally, all data from the STM32MP157 board are transmitted to the Android device, where inference and training of machine learning models are executed. The communication is facilitated by the MQTT protocol, with the broker hosted on the Android device. The data transfer is conducted via an external WiFi connection, specifically established for the H-F application to enforce privacy. As fallback options, the two devices can alternatively connect via an Ethernet cable or a WiFi hotspot generated by the STM32MP157 board. 

In the context of data streaming, particular attention is given to the time synchronization of heterogeneous data, given that signal flows originate from multiple devices but require a common time reference for accurate timestamping. Examples in the literature show that precise time synchronization of input signals is essential to improve the accuracy of data fusion and ML algorithms \cite{gao2023benchmarking} \cite{heydarian2023rwisdm}. 
In \cite{time-sync-framework}, a simple and lightweight framework for synchronizing Bluetooth Low Energy-based sensors is proposed. In this work, we implemented a similar mechanism, despite the differences in the Bluetooth protocol and the higher-level access constraints imposed by the Bitbrain SDK. Specifically, the synchronization process follows these steps:

\begin{enumerate}
    \item The first packet received from the sensor is timestamped by the STM32MP157 board using its internal clock, considering the average round-trip time provided by Bitbrain. 
    \item The following samples are timestamped incrementally, starting from the initial reference instant, adding the sampling period at which the sensor is configured.
    \item Every 10 seconds, the timestamping from point 1 is repeated. 
\end{enumerate}

This synchronization approach effectively mitigates time drift over prolonged sampling sessions, ensuring the accuracy and reliability of the processed signals.

\section{Mental State prediction} \label{sec: mental}

\subsection{Physiological data} \label{sec: data}
The mental state evaluation model is based on physiological data, particularly EEG, ECG, EDA, EMG, and RESP signals. EDA, ECG, RESP, and EMG data are acquired using the Bitbrain Versatile Bio device, and the EEG signals are acquired using the Bitbrain Diadem. These signals are streamed and synchronized using the sensors SDK with the STM32MP157 board as described in \ref{sec: sw}. Biophysical signals are segmented into 60-second windows, while the EEG signals are divided into 5-second windows, resulting in 12 EEG windows for each biophysical one. 

Each windowed signal is passed through its pre-processing pipeline. The ECG signals were filtered using a combination of a band-pass filter (with a frequency range from 0.05 to 40 Hz) and a Savitzky–Golay filter. Electromyography signals were filtered using a band-pass filter with a frequency range from 10 to 500 Hz coupled with a detrending algorithm. The Electrodermal activity signal was filtered using a low-pass filter with a cut-off frequency of 10 Hz coupled with a convolutional signal smoothing. Then, the signal is down-sampled at 100 Hz and divided into phasic and tonic components using the algorithm presented in \cite{grecoCvxEDAConvexOptimization2016}. Respiration signals were filtered using a second-order band-pass filter with a frequency range from 0.03 to 5 Hz. Electroencephalogram signals were processed using two filters: a second-order band-pass filter with a frequency range from 0.5 to 35 Hz and a band-stop filter from 49 to 51 Hz to remove the amplifier noise.

After pre-processing, multiple features were extracted from the biophysical signal windows. $250$ features are obtained from ECG, EDA, and EMG data. ECG features include Heart Rate Variability (HRV) metrics, obtained from time-domain (\textit{i.e.}, heart rate - HR, standard deviation of NN intervals - SDNN), frequency-domain (\textit{i.e.}, absolute power of the low-frequency band - LF), and complexity metrics (\textit{i.e.}, refined composite multiscale entropy - RCMSE). We evaluated the signal energy and Discrete Wavelet Transform (DWT) coefficients' statistics from the processed EMG signal. EDA features include statistical metrics of the Skin Conductance Response (SCR) and time-frequency domain metrics such as Mel-frequency cepstral coefficients (MFCC) and related statistics and DWT Coefficients. The respiration signal also includes measurements like the rate and different measurements of its variability. Concerning the EEG signals, we compute seven features for each of the 12 channels and each window. We evaluate the power in the frequency bands (Gamma (30-80 Hz), Beta (13-30 Hz), Alpha (8-13 Hz), Theta (4-8 Hz), and Delta (1-4 Hz)) and two entropy measures: Differential entropy (DiffEn) and Sample Entropy (SampEn), with a delay of $8$ samples and dimension of $2$.

\subsection{Ground Truth}
Throughout the experiment, ground truth data were collected by administering the Stress Trait Anxiety Inventory-Y1 (STAI-Y1) \cite{stai} questionnaire. It consists of $20$ questions that measure the subjective feeling of apprehension and worry and is often used as a stress measurement. 
The obtained STAI scores are normalized (\textit{min-max}) by subject and coupled with the physiological data obtained during the task. The normalized STAI score is the regression output of the proposed model.

\subsection{Model Architecture}
To detect stress levels from the given evaluated features, we developed a multimodal model, which takes as input a tuple of data $(x_e, x_p)$. $x_e$ represents the feature matrix of the EEG signal, with shape $(b, n_w, n_f, n_c)$, where $n_w$ is the number of windows, $n_f$ is the number of features, and $n_c$ is the number of EEG channels. $x_p$ represents the feature vector of the other biophysical signal, with shape $(b, n_p)$, where $n_p$ is the number of features obtained from the 60-second window. 

\begin{figure}
    \centering
    \includegraphics[width=0.99\linewidth]{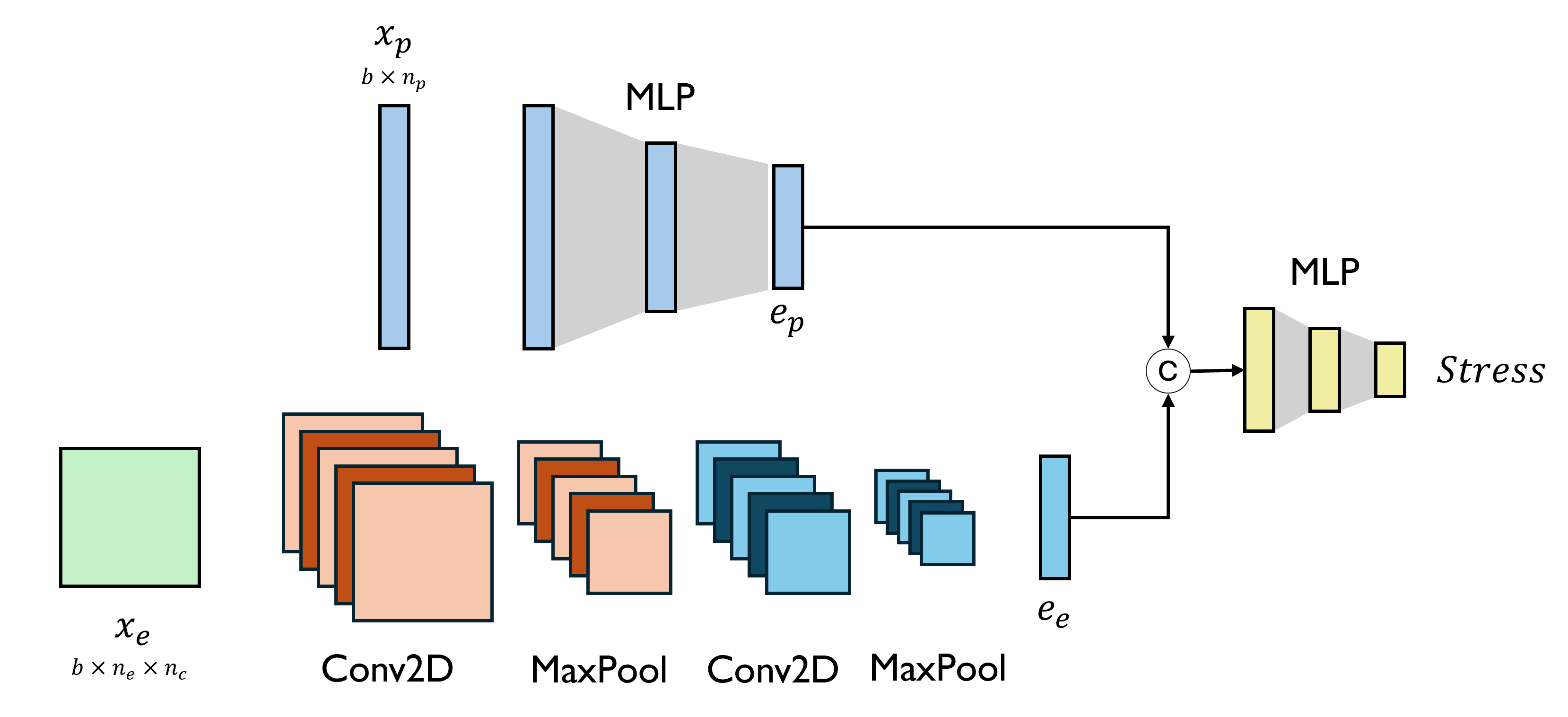}
    \caption{Schematic representation of the Multimodal model architecture.}
    \label{fig: model}
\end{figure}

The model consists of two main backbones to process physiological and EEG features independently. The EEG backbone consists of 2 convolution blocks, each consisting of a Conv2D layer with ReLU activation, followed by a MaxPool2D. The convolution channels are $(32, 32)$, while the kernel shapes are $(3, 3)$ and the pooling layers kernel are $(2, 2)$, with the $(2, 2)$ strides. The biophysical backbone consists of a Multilayer perception with two hidden layers with $64$ and $128$ neurons. The obtained representations are then concatenated, using a late fusion approach, and fed to an MLP regressor with one hidden layer with $64$ neurons. A schematic representation of the model can be seen in Fig. \ref{fig: model}.

\section{Participants}
In total, $33$ subjects participated in the data collection. The sample mean age is {$31.94 \pm 11.95$}. $27$ subjects were male and $6$ were female. Most subjects were invited from the author's research facility, while the others accepted an external invitation. Participant background varies from undergraduate engineering students to researchers, including professionals in other fields. All participants provided informed consent before participating in the study. The consent process included information about the purpose of the experiment, how the data would be collected, stored, and used, and relevant ethical considerations related to the handling of psychophysiological data. The procedures complied with the ethical guidelines of the hosting institution.

\section{Experimental Setup} \label{sec: exp}

\begin{figure}
    \centering
    \includegraphics[width=\linewidth, keepaspectratio]{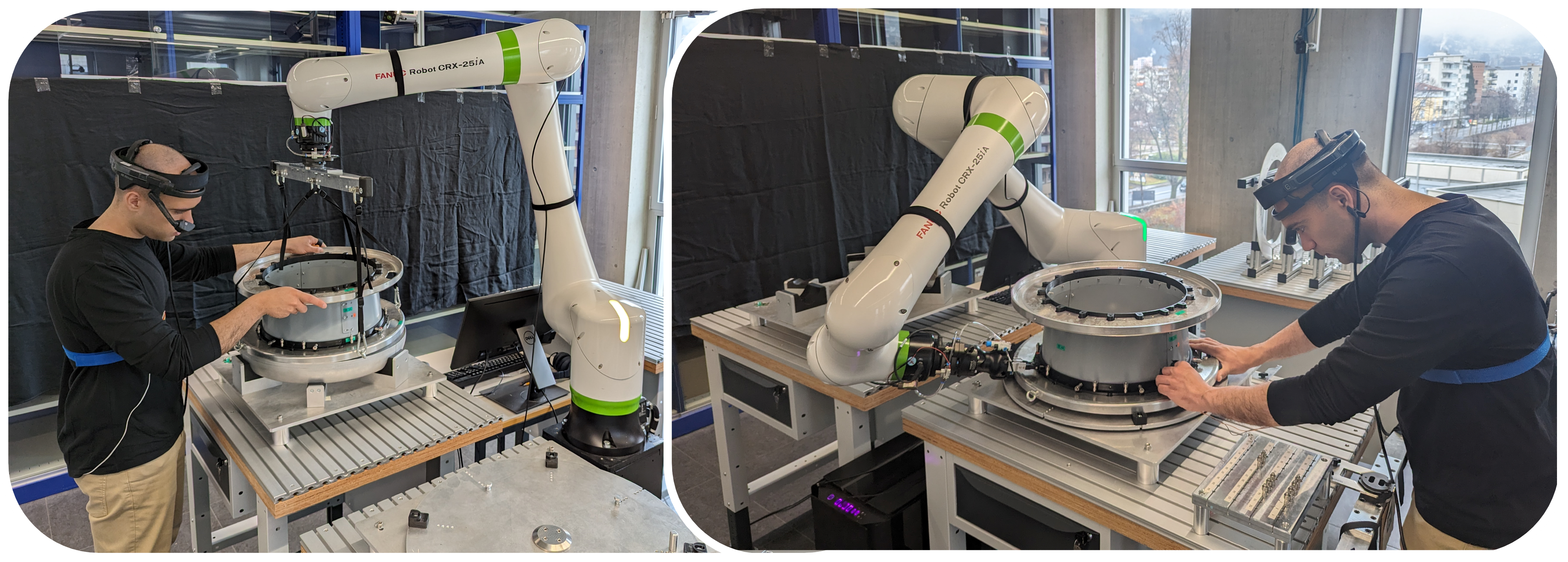}
    \caption{A human operator collaborates with a Fanuc CRX-25iAL cobot to assemble an airplane engine. Physiological data were collected to analyze the human state during the assembly process.}
    \label{fig: cell}
\end{figure}

Physiological data were collected while performing a collaborative assembly of an airplane engine nacelle. The task can be decomposed into three main steps: the cooperative transport of components on the assembly table; the insertion of multiple fasteners performed by both human operator and robot; and the collaborative transport of the sub-assembly components. The work cell is equipped with a Fanuc CRX-25iAL\footnote{https://crx.fanucamerica.com/crx-25ia} cobot. An example representation of the task and the robotic cell can be seen in Fig. \ref{fig: cell}.
To gather data, we asked participants to perform the assembly $5$ times. After each assembly, the operator was asked to answer the STAI-Y1 questionnaire. We perform subject-specific min-max normalization on the questionnaire scores and the evaluated features. These data, including the corresponding labels obtained from the questionnaire, are processed and stored anonymously locally inside the H-F device, which is password protected. Consequently, these labeled samples can be used for distributed training.

Given the computational requirements of on-device training, the clients were deployed on Android devices with high computational power. We chose a commercially available smartphone, the Nubia REDMAGIC 8 Pro. The high-performance device communicates with the STM32MP157 board, mentioned in Section \ref{sec: sw}, through the MQTT protocol to receive the physiological data. Moreover, the Android device communicates with the server using Flower’s communication framework, which relies on Google Remote Procedure Call (gRPC) which uses Protocol Buffers (Protobuf) for compact message exchange and supports bidirectional streaming for handling tasks like sending model updates and receiving training results. This bidirectional communication is essential in federated learning scenarios, as it allows clients to send their locally trained model weights to the server and receive updated global model weights in return. An Android client was developed to interact with a server deployed on R-F. The main working principle is the same. However, personal data are collected, processed, and stored on the H-F device using an anonymous user-specific ID. The strategy used by the server to update the global model is FedAvgAndroid, which is a declination of the FedAvg strategy specifically designed for Android-based clients, such as the H-F. This strategy accommodates the constraints and characteristics of Android devices, such as limited computational resources, intermittent connectivity, and reliance on mobile frameworks. In this scenario, the server application runs on R-F and communicates with the available H-F. The models of the logged-in users are trained on-device before sending the updated weights to the server, and model checkpoints are saved on the device using the user ID. The Android Flower client was developed using the model described before and converted to TensorFlow lite.

\section{Results} \label{sec: res}
\begin{figure}[b]
    \centering
    \includegraphics[width=\linewidth]{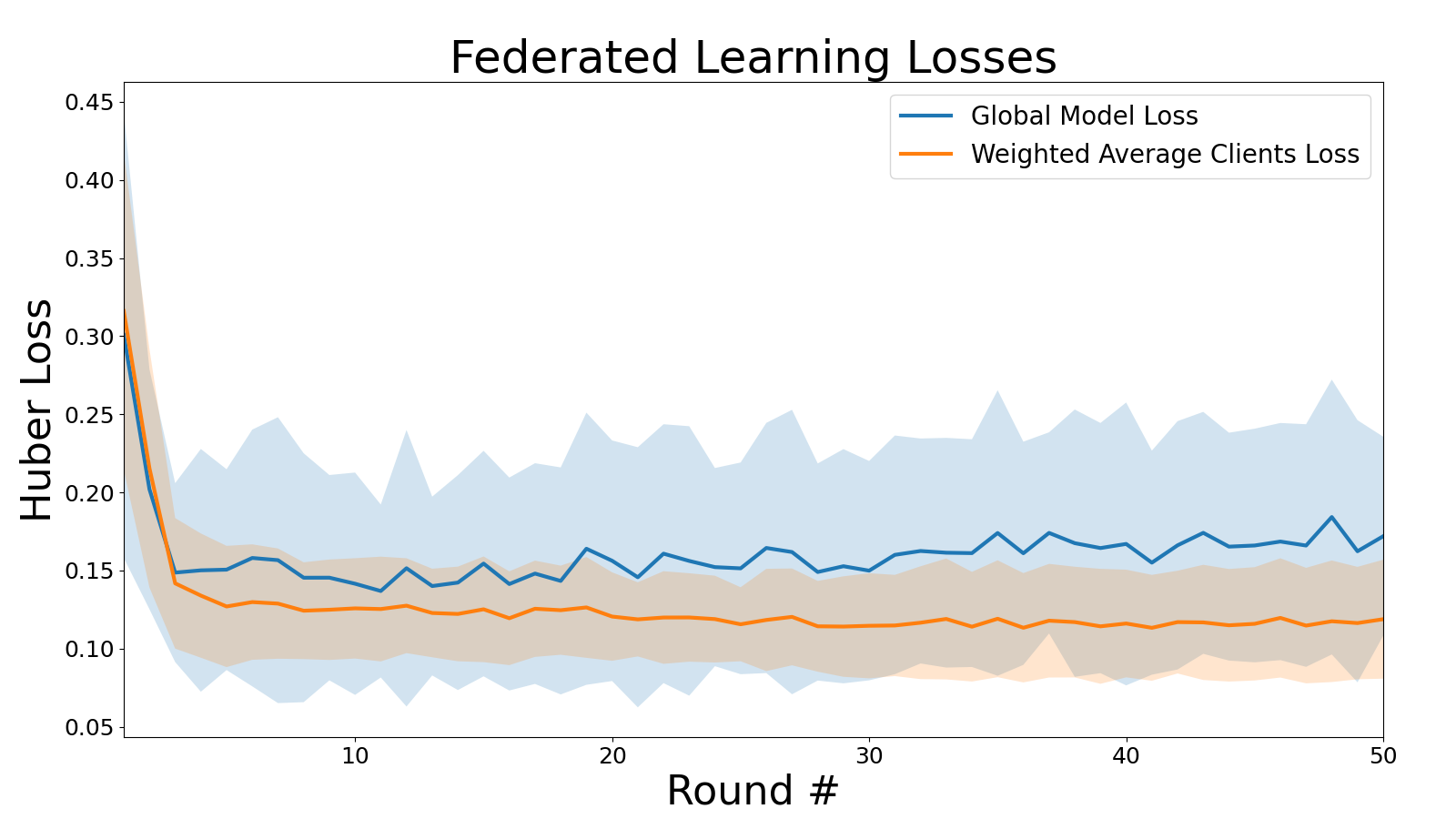}
    \caption{Loss obtained from the weighted average of the client's loss (orange) and loss obtained using the global model evaluated over the test subject. The results are shown as the round's average and standard deviation over each test subject.}
    \label{fig: loss}
\end{figure}

To evaluate our approach, we first compare the performance of a model trained with a distributed approach against a centralized model. To achieve this, we performed \textit{Leave-One-Subject-Out} (LOSO) validation on both training approaches.
The model trained with a centralized approach was trained for $500$ epochs with the AdamW optimizer, with a learning rate of $1e-4$ and early stopping in case of overfitting.
For each test subject, the FL model was trained for 50 rounds on all available clients. During each training round, the server collected data from a minimum of 8 random clients among all the available ones. Then, the obtained aggregated global model was tested on the left-out subject. The global model was initialized using random weights. To simulate the availability of different users, each client was activated and deactivated after a random amount of time during the global model training rounds. Each client model was trained using AdamW with a learning rate of $1e-4$ for $5$ local epochs. For both training methods, we employed the Huber loss:

\begin{equation}
    L_{\delta}(a) =
        \begin{cases} 
        \frac{1}{2} (y - \hat{y})^2, & \text{for } |y - \hat{y}| \leq \delta \\
        \delta (|y - \hat{y}| - \frac{1}{2} \delta), & \text{for } |y - \hat{y}| > \delta
        \end{cases}
\label{eq: huber}
\end{equation}

where $\delta$ is a threshold that determines the transition between quadratic and linear loss and has a chosen value of $0.05$.
In Tab. \ref{tab: comparison} we report the RMSE of the best epoch/round for each of the two models. 

To assess the convergence of our approach, for each subject in the LOSO validation, we analyze the convergence of the global model over the specific test subject, and we compute the RMSE weighted average of the clients' models as follows:

\begin{equation}
    \text{RMSE}_{\text{weighted}} = \frac{\sum_{i=1}^{N} n_i \cdot \text{RMSE}_i}{\sum_{i=1}^{N} n_i}
\end{equation}

where $n_i$ represents the number of samples available on each client. As can be seen in Fig. \ref{fig: loss}, the global model loss, evaluated over the test subject, decreases after each training round and converges to the lower bound. It also shows that the client models' weighted average improves their performance after each round of receiving a global model update. This validation underlines the effectiveness of the federated learning approach in maintaining high model accuracy while preserving data privacy.

\begin{table}[b]
\centering
\footnotesize
\renewcommand{\arraystretch}{1.5}
\begin{tabular}{ c | c | c }
    Validation Metric & Centralized & Distributed (\textit{FedAvg}) \\ [0.5ex]
    \hline \hline
    RMSE & $0.38 \pm 0.076 $ & $0.26 \pm 0.05 $ 
\vspace{1.5em}
\end{tabular}
\caption{Comparison between a centralized training model and the same model trained in a Federated Learning framework.}
\label{tab: comparison}
\end{table}

During FL training, each client's fine-tuned model was stored locally after every round for the respective subject. In this way, we can introduce the personalization of the mental state model. To highlight the improvements of the personalized solution over the global model, we compared the results obtained with global models and the personalized model on that particular subject. The models taken for comparison are all from the final round of the FL training. In Tab. \ref{tab: personalized} we compared the RMSE obtained as the mean performance of the global model, which includes the specific subject as one of the clients, and the results obtained from the model personalized on that specific subject. As expected, the personalized model improves the performance on that particular subject, without ever sharing the subject's data.

\begin{table}[t]
\centering
\footnotesize
\renewcommand{\arraystretch}{1.5}
\begin{tabular}{ c | c | c }
    Validation Metric & Personalized & Global model \\ [0.5ex]
    \hline \hline
    RMSE & $0.25 \pm 0.06 $  & $0.34 \pm 0.07 $ 
\vspace{1.5em}
\end{tabular}
\caption{Comparison between the global models and the personalized model on each client.}
\label{tab: personalized}
\end{table}

\section{Discussion} \label{sec: disc}
The results indicate that the proposed personalized mental state prediction approach has significant potential for seamless deployment in industrial settings,  maintaining model performance while preserving data privacy. Compared to centralized training, the distributed FL approach demonstrates improved robustness, particularly in scenarios with varying user characteristics. The personalized models further improved stress prediction accuracy, highlighting the value of personalization in human-robot collaboration (HRC). This personalized approach enables the robot to better interpret an operator’s mental state, potentially improving collaboration fluency and reducing cognitive load.

Despite these advantages, certain limitations must be addressed. The dependency on high-performance edge devices for on-device training may limit scalability in real-world industrial settings. Additionally, variations in data quality across clients can impact FL convergence, requiring further investigation into adaptive aggregation and personalization strategies. Furthermore, industries face strict regulations regarding the collection and processing of personal data. Continuously updating models requires ongoing transmission of sensitive user data, even in aggregated or anonymized form, which can raise legal and ethical issues. 
When issued as a service, the proposed framework will have to deal with data usage policies, finding an agreement with workers, adopting company and service provider. In fact, users could negate the consent to model personalization or to weight aggregation by FL. A fair solution could be to establish an agreement where, if users want to benefit from model personalization, they have to accept weight sharing for improving the baseline model. If they don't agree, they can still rely on the globally trained model. Although the global model provides a general understanding of human states based on aggregated data, it lacks the specificity needed to fully capture an individual’s unique physiological and behavioral patterns. As a result, the robot adaptation may be less precise, potentially limiting the overall effectiveness of the interaction.
When data are only sporadically available, ruling out continuous personalization, the framework can still support a middle-ground alternative. From the perspective of the FL framework operation dynamics, research on personalized FL typically assumes that model updates will occur when new data are available. In light of this,  we can argue that a snapshot or a frozen version of a personalized model can adequately meet a user's needs for a limited time. By freezing the model after initial fine-tuning, companies can minimize data collection and reduce the regulatory burden associated with continuous learning. Periodic re-personalization could mitigate the risks associated with data drift and reduced model performance in highly dynamic contexts.
This trade-off can potentially result in smoother collaboration and improved task efficiency in real-world manufacturing settings, as the robot can better align its behavior with the human colleague’s needs. 
In summary, while continuous personalization through federated learning demonstrates high potential for enhancing human-robot collaboration, the trade-offs in terms of privacy, integration complexity, operational costs, and safety call industrial practitioners to opt for strategies carefully tailored to their specific organizational context. 

\section{Conclusion} \label{sec: concl}
This study proposed a federated learning framework for personalized mental state evaluation in human-robot collaboration, ensuring data privacy while enabling individualized stress predictions. The results demonstrate that FL not only preserves user privacy but also outperforms traditional centralized models in predicting stress states. Furthermore, personalization significantly enhances prediction accuracy for individual operators, reinforcing the importance of adaptive models in industrial HRC.

Future research should focus on optimizing FL strategies to improve robustness across diverse industrial environments. Additionally, integrating other physiological and behavioral modalities, such as facial expressions or speech characteristics, could further enhance stress detection accuracy. Ultimately, this work contributes to advancing human-centered robotics by enabling real-time, personalized, and privacy-preserving adaptation to an operator’s mental state.

\section{Match \& Contribution}
This contribution aligns closely with the theme of the ICE IEEE 2025 conference on “AI-driven Industrial Transformation: Digital Leadership in Technology, Engineering, Innovation \& Entrepreneurship.” The paper introduces a personalized, privacy-preserving framework for mental state evaluation in Human-Robot Collaboration (HRC) using Federated Learning (FL). By addressing the challenges of data privacy and personalization, the research contributes practical frameworks that leverage multimodal physiological data to enable real-time robot behavior adaptation. The proposed system is implemented in an industrial setting and demonstrates the feasibility of distributed on-device training for stress prediction, showcasing how emerging AI technologies can create tangible value by improving worker well-being and collaboration fluency. This work directly addresses the management of emerging technologies, analyzes key implementation challenges in edge AI and HRC, and exemplifies value creation through personalization and human-centered automation.

\section{Acknowledgements}
This research has been funded by the Horizon Europe project Fluently (Grant ID: 101058680). The authors would like to thank all participants involved in the data collection process and the technical staff supporting the deployment of the experimental setup.



\end{document}